# Categorizing Online Shopping Behavior from Cosmetics to Electronics: An Analytical Framework

Sohini Roychowdhury[1], Wenxi Li[2], Ebrahim Alareqi[3], Akhilesh Pandita[2], Ao Liu[3], Joakim Soderberg[3]
[1]Fourthbrain.ai, Palo Alto, CA,
[2]University of California, Berkeley, CA.
[3]Volvo Cars Technology, CA

***Abstract*—A success factor for modern companies in the age of Digital Marketing is to understand how customers think and behave based on their online shopping patterns. While the conventional method of gathering consumer insights through questionnaires and surveys still form the bases of descriptive analytics for market intelligence units, we propose a machine learning framework to automate this process. In this paper we present a modular consumer data analysis platform that processes session level interaction records between users and products to predict session level, user journey level and customer behavior specific patterns leading towards purchase events. We explore the computational framework and provide test results on two Big data sets – cosmetics and consumer electronics of size 2GB and 15GB, respectively. The proposed system achieves 97-99% classification accuracy and recall for user-journey level purchase predictions and categorizes "buying" behavior into 5 clusters with increasing purchase ratios for both data sets. Thus, the proposed framework is extendable to other large e-commerce data sets to obtain automated purchase predictions and descriptive consumer insights.**

## I. Introduction

Recent times and safety measures have necessitated availability of e-Commerce platforms for almost all products including a seamless shopping experience. With the current paradigm shift in digital shopping experience brought about by social distancing during a pandemic, there is a critical need for scalable analytical frameworks. Such frameworks must automatically analyze virtual shopping experiences to aid shopping personalization, inventory management, and marketing for customers and manufacturers, respectively. The primary reasons for disparity in aggregated virtual shopping processes across products and platforms include variations in product cost, delivery wait times and ease of shopping platform ease for usage [1-2]. In this work, we analyze online shopping history from customers using a consistent framework that can scale across products and platforms to identify patterns and trends over time, which can then be representative of specific shopping "behaviors". The proposed system is designed to predict customer conversion from browsing to purchasing at a session level and at user journey levels [3-6], as shown in Fig. 1.

Our goal here is to create a common workflow that performs feature engineering, feature selection followed by predictive modeling for user-product interactions (as user journeys). The system also predicts an upcoming purchase event at a session level using sequence modeling. Finally, we categorize customer purchasing behavior into five distinct categories based on the product viewing, carting and purchasing behaviors that aids customer insights to guide business/marketing intelligence. Such an analytical and scalable framework for predicting purchase events based on user-product interaction levels, session levels and customer category levels has not been developed till date.

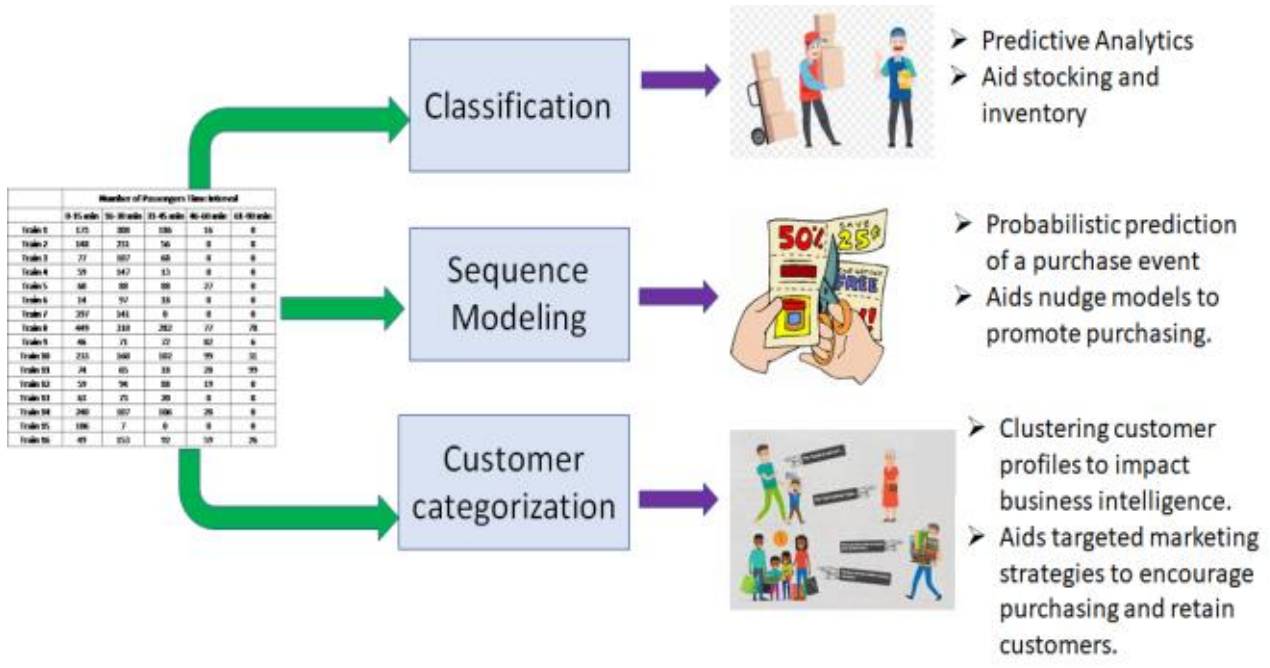

Fig. 1. Proposed modular workflows for analyzing online shopping data for predictive classification of purchase events, sequence modeling for probabilistic prediction of a purchase and customer categorization to direct promotional events to maximize purchasing outcomes.

When customers login to a shopping website, they accept cookies to establish the session. The session-ID, client-ID combination can then be used to uniquely log information regarding product-level browsing, additions to cart, removals from cart and purchase etc. The session level data can then be consolidated to create user-journeys, as shown in Fig. 2, to then analyze the propensity for sale per-client for each product-type interaction. Existing works so far have analyzed product demands differently for product price variations such as the variation in price and demand for cosmetics vs. electronic items [1-2]. In this work, we utilize a user-product journey of events and session level events to make the following three major contributions.

- Scalable feature engineering with automated feature selection capability to identify the most relevant features that lead to a purchase event. The optimal feature sets are useful to achieve 97-99% classification accuracy and recall for purchase events at user journey level for the cosmetics and electronics data sets, given that the data sets represent data imbalance with 12.06/1% purchase ratio (PR), respectively.

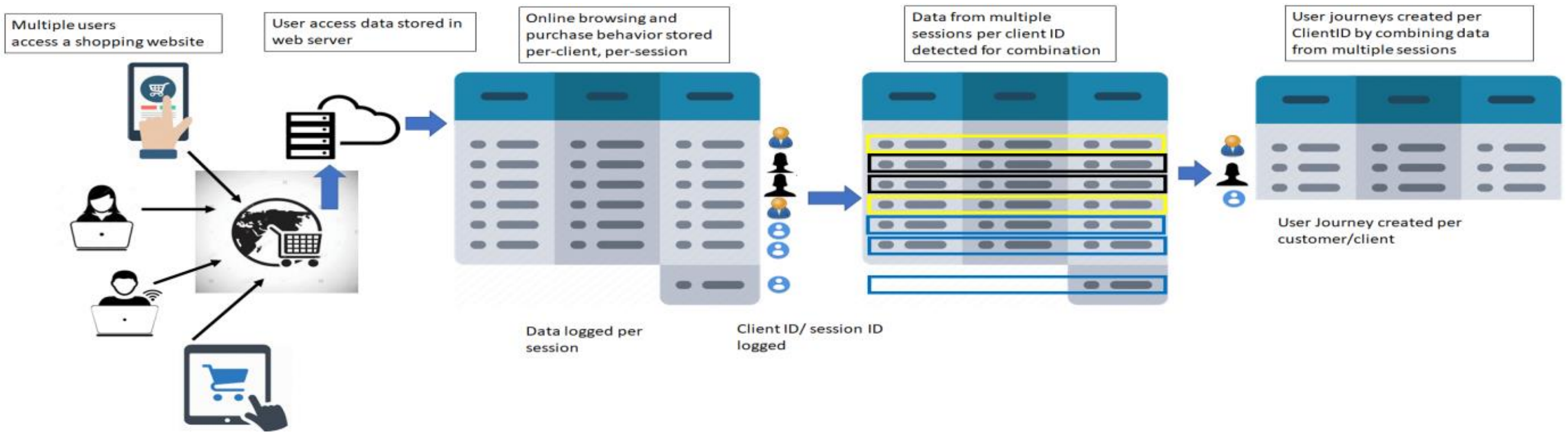

Fig. 2. Example of data flow to create User Journeys from several purchase sessions. As multiple customers access the web-based shopping portal, records corresponding to the events (view, cart, remove, purchase) are time stamped and recorded for each session. Data from sessions are then combined per unique UserID to create User-journeys.

- Sequence modeling to predict purchasing events at session levels are useful to design customer *nudge* models for product promotion. We achieve 91-97% accuracy and 43-77% recall in predicting a purchase in the following session. This is a significant improvement over baseline sequence models [3].
- Customer behavior - categorizing by clustering in the t-SNE manifold space followed by feature level analysis for discriminating user journeys with purchase events from the non-purchase journeys, respectively. We identify 5 categories of customer behavior that represent upto 3-8 times increase in PRs based on the interaction levels of users. Based on this analysis, we can separate *New Shopper* behavior from *Returning Decisive Shopper* behaviors that can then inform market intelligence accordingly to propose recommendations aimed to maximize purchasing from each category, individually.

## II. RELATED WORK

Predictive modeling (classification) and sequence modeling for e-commerce session level data has been analyzed in several works for far. For instance, the work in [4] applied the time-stamps of clicks in a clicking stream per session to model the buying patterns using bidirectional LSTM models. This work reported that the outcome was comparable to using feature engineering and classification. Further, in [4], relative features were aggregated and processed with traditional machine learning and deep learning algorithms such as gradient boosting regression and deep neural networks. In this work we implement feature-level aggregation and predictive models but apply it to a variety of data streams.

Beside session level predictions, there is a need to predict repeat customers as in [5] and their tendency to return and finish their orders. This implies that an order can span across multiple sessions and needs data aggregation. This work is motivated by the same principle as we analyze customer conversions at user-journey and session-levels towards purchase events using engineered features that are scalable across product price and demand variations. Data balancing as in [6] is important for classification at user journey level using k-nn, logistic regression models. Further, automated feature selection based on feature importance and output metrics significant to application are identified as in [7].

## III. MATERIALS AND METHODS

In this section we describe the data sets under analysis, the methodical and computational frameworks for the proposed system.

### *A. Data Description*

Here, we analyze two public datasets acquired from Kaggle for a comprehensive experiment series, including eCommerce Events History in Cosmetics Shop [1], as well as eCommerce behavior data from multi category store [2]. For the second dataset, electronics data is filtered to keep the study market specific and make it easy to compare different consumer behavior patterns on different types of products. From the raw data, we engineer features at session-level and user-journey level in Section IV (details in Supplementary Materials for readability). Both of the datasets have the same columns including user ID, event type (cart, view, remove from cart and purchase), time stamp of the event, as well as product metadata such as product category, brand, price, userID and user session ID as shown in Fig 3.

**Original Data: Based on every single event**

| | event_time | event_type | product_id | category_id | category_code | brand | price | user_id | user_session |
|---|---|---|---|---|---|---|---|---|---|
| 0 | 2019-12-01 00:00:00 UTC | remove_from_cart | 5712790 | 1487580005268456287 | NaN | f.o.x | 6.27 | 576802932 | 51d85cb0-897f-48d2-918b-ad63965c12dc |
| 1 | 2019-12-01 00:00:00 UTC | view | 5764655 | 1487580005411062629 | NaN | cnd | 29.05 | 412120092 | 8adff31e-2051-4894-9758-224bfa8aec18 |
| 2 | 2019-12-01 00:00:02 UTC | cart | 4958 | 1487580009471148064 | NaN | runail | 1.19 | 494077766 | c99a50e8-2fac-4c4d-89ec-41c05f114554 |

**Transformed Data: Based on user journey**

| product_id | user_id | # Clicks | Sales | # Sessions | last_time | max_price | min_price | year | month | weekday | hour | timeOfDay | # Cart | # View | insession_cart |
|---|---|---|---|---|---|---|---|---|---|---|---|---|---|---|---|
| 1000365 | 525766088 | 1 | 0.0 | 1 | 2019-11-05 16:55:19 | 1029.09 | 1029.09 | 2019 | 11 | Tuesday | 16 | Afternoon | 0 | 1 | 0.0 |
| 1000978 | 424872790 | 1 | 16.0 | 1 | 2019-11-21 11:07:30 | 359.47 | 359.47 | 2019 | 11 | Thursday | 11 | Morning | 0 | 1 | 0.0 |
| 1000978 | 442462854 | 8 | 16.0 | 4 | 2019-11-21 09:13:56 | 359.47 | 300.91 | 2019 | 11 | Thursday | 9 | Morning | 0 | 8 | 0.0 |
| 1000978 | 460216566 | 1 | 16.0 | 1 | 2019-11-21 16:00:46 | 359.47 | 359.47 | 2019 | 11 | Thursday | 16 | Afternoon | 0 | 1 | 0.0 |
| 1000978 | 460263198 | 2 | 16.0 | 1 | 2019-11-13 [illegible] | 300.91 | 300.91 | 2019 | 11 | Wednesday | 7 | EarlyMorning | 0 | 2 | 0.0 |

Fig. 3. Example of raw data from the cosmetics market and corresponding engineered features.

### B. *User Journey Analysis: Baseline*

A user's interactions with a given product can be modeled as a user journey. An ideal user journey starts at the view event, progresses to the cart event, and then ends at the purchase event. However, there are multiple permutations for this journey. For example, some users can directly go to purchase post viewing an item while others may come into the pipeline through a third-party source. Also, some users may directly purchase an item while others may view and cart the item multiple times then remove it while making a purchase in a few cases and no purchases in others. In the absence of a data model, a baseline method is to plot the different purchasing paths (consisting of session level web-page visits on a time scale) and finding the relative frequency of each path that leads to a purchase event as shown in Fig. 4.

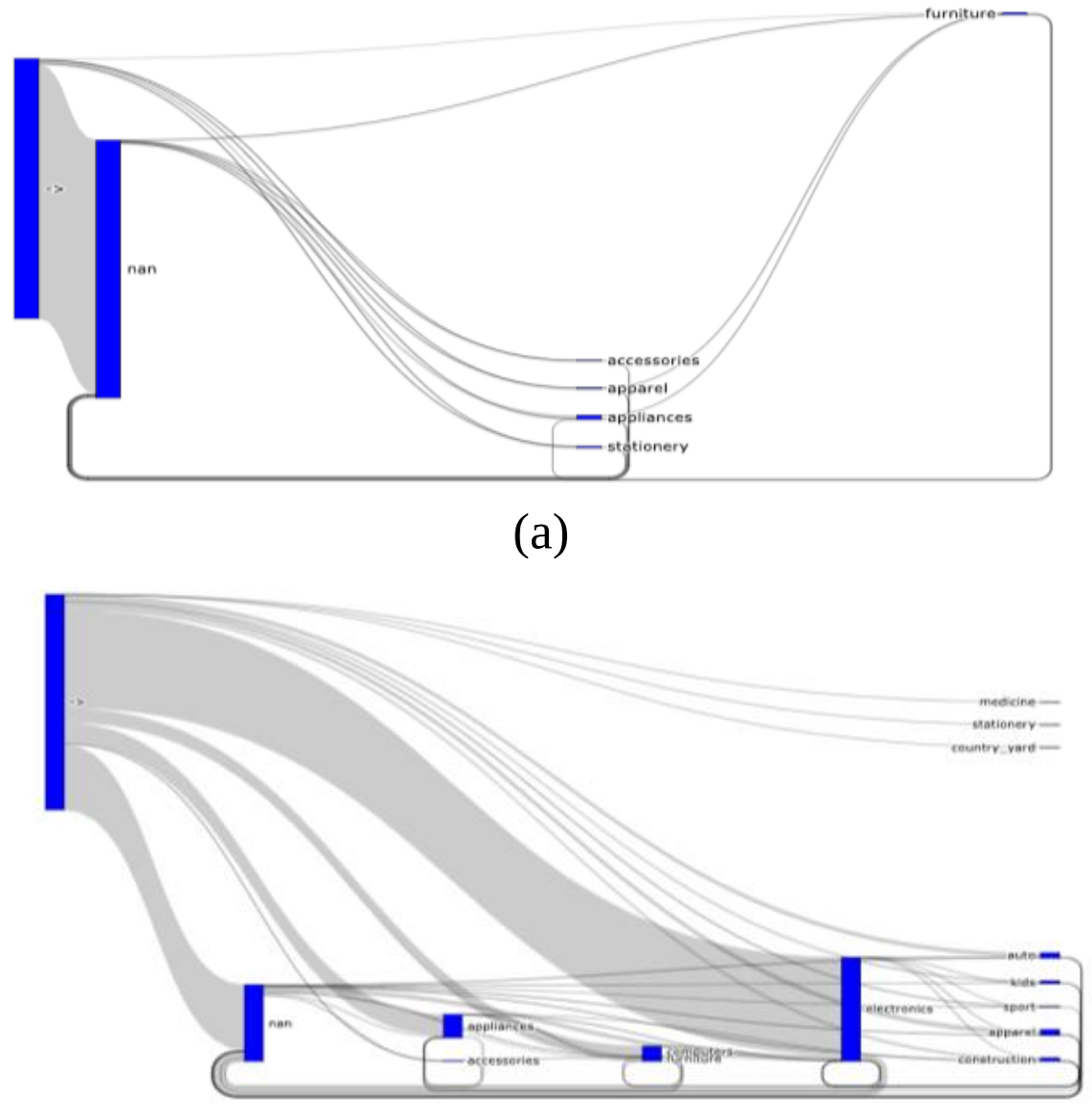

(a)

(b)

Fig. 4. User Journey created using time-stamped customer product interaction data for (a) Cosmetics data set (b) Electronics data set, respectively.

In Fig. 4(a), the Sankey graph depicts the top 100 most frequent page sequences grouped by userId. The average sequence length is 5.3 with several loops back to a root page. The graph is simplified by showing only top-level pages categories. However, this baseline method fails to identify key factors/features that lead to a purchase event, thereby necessitating a modular analytical framework as the one proposed in this work.

### C. *Methods and Formulations*

Each raw data set, denoted by $X_0$comprises of all metadata fields per-session apart from the event type, which is denoted by $Y_0$. From $X_0$, features are engineered per (product ID, UserID) pair, denoted by $X_S \epsilon R^{mXd}$and $X_J \epsilon R^{nXf}$for features extracted at session- level and user journey levels, respectively. The session based features $X_s$ have the dimensionality [4535940x 41] and [12182304x34] for the cosmetics and electronics datasets, respectively, while the user journey based dataset have the dimensionality [10156200x33] and [22059716x27] for the cosmetics and electronics datasets, respectively, (excluding the outcome label). The outcomes $Y_S$ and $Y_J$ represent a binary event labels, respectively, where $Y_{(K,i)} = 1$ , for $K = \{S, J\}$ signifies a session record $i$ to have concluded in a purchase event, while $Y_{(K,i)} = 0$ signifies product view, added or removed to cart but not purchased. Following the feature selection process, the top retained features at session and user journey level are $X_{S'} \epsilon R^{mXd'}$and $X_{J'} \epsilon R^{nXf'}$, respectively, where $d' \leq d$, $f' \leq f$. Data and label sets $\{X_{S'}, Y_S\}$and $\{X_{J'}, Y_J\}$ are then subjected to classification. For the user journey data, we use data models such as logistic regression and K-Nearest Neighbors (KNN) models with balanced datasets to counteract the data imbalance problem [8]. For session-level data, we apply the Long-Short Term Memory (LSTM) models varying in network structure (model parameters) to learn patterns at a session-sequence level.

The number of class 0 and class 1 events that are classified correctly are referred to as true negatives $(tn)$ and true positives $(tp)$, respectively. The events that were actually 1 but misclassified as 0 are false negatives $(fn)$, while events that were actually 0 but misclassified as 1 are false positives $(fp)$. The output metrics evaluated thereafter are as follows.

$$Precision = \frac{tp}{tp + fp} \quad (1)$$

$$Recall = \frac{tp}{tp + fn} \quad (2)$$

$$Accuracy = \frac{tp + tn}{tp + tn + fp + fn} \quad (3)$$

The data sets under analysis demonstrate high data imbalance with the ratio of non-purchase (Y=0) to purchase (Y=1) being nearly 7:1 in the cosmetics dataset and 35:1 in the electronics dataset when aggregated at a user-journey level. The same ratio becomes 28:1 in the cosmetics dataset and 16:1 in the electronics dataset when aggregated at a session level. Besides, it is more important not to miss a customer conversion event resulting in a purchase (to ensure enough stock availability for customers) than to over predict a conversion event (that can cause excess inventory but prevent shortage). Thus, for the classifiers, $fn$ should be reduced and *Recall* metric in (2) should be maximized rather than the *Accuracy* metric in (3) that considers $fn$ and $fp$to have equal weightage.

As a final step to categorize user-behavior, we use the journey data $X_J$,and perform unsupervised clustering in the t-SNE probabilistic plane, to identify *similar buying* patterns in each cluster. The t-SNE plane preserves local similarities between sample points, which promotes clustering patterns using k-means or DBSCAN methods.

The percentage of samples belonging to each cluster (Rep) and the PR in each cluster identify distinct buying behaviors. These behaviors are based on the length of interaction time, numbers of sessions and time spent in carting, viewing and removing actions per session and in the overall journey. These buying behaviors are then analyzed at feature levels to provide customer insights for market intelligence.

### *D. Computational Framework*

Both datasets under analysis contain several million records, which poses a scalability issue from the processing perspective. For our system developed in Python, working with *Pandas* library alone can lead to significant storage issues due to memory expansion during a complex transformation/aggregation operation or even during data import. For our proposed system, we have leveraged several techniques described below to reduce computational complexity and enable large data set processing.

i. *Chunking the data* - The simplest way to solve the file reading problem is to separate the data in several chunks, each of which is readable in Python. After aggregating each chunk, abundant information is deleted and the data size shrinks at an aggregated level.

ii. *Using Pivot table and Unstack* - For creating new features that require data aggregation, *groupby* command is typically followed by *apply* or *map*. However, the *apply* function is not only extremely memory intensive when dealing with large datasets having 1 million or higher number of records (at times leading to kernel crashes), it is complex in application as well.

iii. *Merging dataframes* - While *pandas.merge* operation is faster than the *join* operation on limited records, as the data size increases, *join* proves to be a faster option. To illustrate the execution time comparison, on a 16 GB memory virtual machine it takes around 9.5 seconds to merge a dataframe of shape [4.5 million samples x23 features] using pandas.merge, but it takes only 5.2 seconds using the *join* operation.

iv. *Using pd.to_datetime inplace of datetimeindex* : To engineer the date-time related features, pandas.DatetimeIndex and pandas.to_datetime produce the same results, however, *pandas.to_datetime* with the input date-time format specified is a faster process.For e.g. for 1 million records, converting a series data to the Timestamp format and then extracting the year from the Timestamp takes around 9 seconds using pandas.DatetimeIndex functions but only 1.8 seconds using pandas.to_datetime function making it approximately 5 times faster.

v. *Google BigQuery*: Typical e-commerce platforms collect both transactional and behavioral data, that become several gigabyte to terabytes worth of data streams daily. Since traditional OLTP databases often face scalability challenges to meet growing data storage needs, BigQuery [9] has become an important data retrieval method. After importing the data with Big Query, some of the trivial queries can be performed first in SQL. It gives an initial aggregation of the raw data, with redundant information removed. The Big Query process shrinks our data from 2GB to 760MB, and 14GB to 5GB.

vi. *Pickling the data*: The best data practices include saving processed dataframes as *pickled* data, which makes storage and retrieval faster than *csv* format. The *hdf5* file format is another option instead of pickle. In our case both options proved to be equally suitable.

## IV. Feature-level Analysis

The first step towards development of predictive data models is feature engineering and optimal feature set selection. Accordingly, Exploratory Data Analysis (EDA) is performed to understand the main characteristics of our datasets and to identify features of importance (see Supplementary Material). The feature sets at user-journey and session levels are described below.

### *A. User-Journey Level Analysis*

From the raw data, we observe 6.22% of the events in the cosmetics dataset are purchase events as compared to only 1.51% of the events in electronics dataset as shown in Fig. 5. It is noteworthy that the purchase event percentage is calculated based on single events and this percentage will be higher after data being aggregated to session level, and even higher after aggregation to user-journey level.

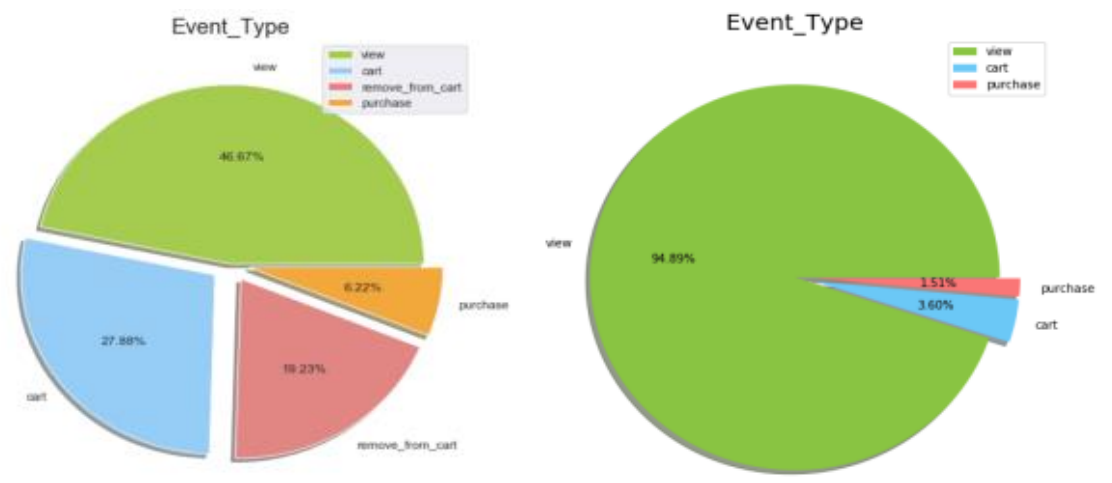

(a) (b)

Fig. 5. Pie Charts of Event Types over (a) Cosmetics Data, (b) Electronics Dataset, respectively.

Several works till date [3-4] have focused on session level classification for purchase events. However, session-level analysis misses the repetitive patterns in customer shopping behaviors wherein the same customer may close the shopping window on one day and return sometime later. In such instances, the number of sessions one customer spends on the same product may have different features contributing to a purchase event when compared to a session level analysis. Thus, we extract features corresponding to user -product interactions over time and perform feature ranking to retain only the top ranked features with highest contributions.

For optimal feature set selection, we rank the features using embedded methods like Random Forest [7] and filter methods like Fisher score [7]. While Random Forest generalizes better and also combines the qualities of filter and wrapper methods, we combine the highly ranked features from both methods as the optimal feature set shown in Table I. Feature ranking and EDA demonstrate that there is little to no variance in the distribution of purchase journeys v/s non-purchase journeys when measured against date time attributes. An example of feature ranking is shown in Fig 6, where we observe that features like the number of events in user journey, total interaction time, number of sessions, number of carts, views and removals have significantly higher weightage than features such as time of day, day of week or month of purchase.

TABLE I. SELECTED JOURNEY FEATURES FOR TWO DATASETS

| | **Dataset** | |
|---|---|---|
| | Cosmetics | ***Electronics*** |
| Selected Features | NumOfEventsInJourney<br>NumSessions<br>interactionTime<br>NumCart<br>NumView<br>NumRemove<br>Price<br>InsessionView<br>InsessionCart<br>InsessionRemove | NumOfEventsInJourney<br>NumSessions<br>interactionTime<br>NumCart<br>NumView<br>Price<br>InsessionView<br>InsessionCart |

## *B. Session-level Analysis*

While user journey analysis gives a chance to predict conversion across multiple sessions, a session level analysis is useful to predict sequence of events and success of a specific user-product interaction. There are two specific instances where session-level analysis is more beneficial than journey level. First, if a user logs in with different devices or skips logging in, multi-session history across devices cannot be retained. In such cases session-level decisions must be made to predict a purchase event. Second, sequence models such as LSTM are capable of learning local and global contextual patterns for purchasing behaviors. Thus, on a session level, data collection and processing becomes easier than storing data across sessions for aggregation, and this allows for prediction of purchase behavior as a sequence of events, that aid marketing *nudge* models, or coupons/deals, that can bring forth a purchase event sooner than predicted. The session level features are shown in Table II.

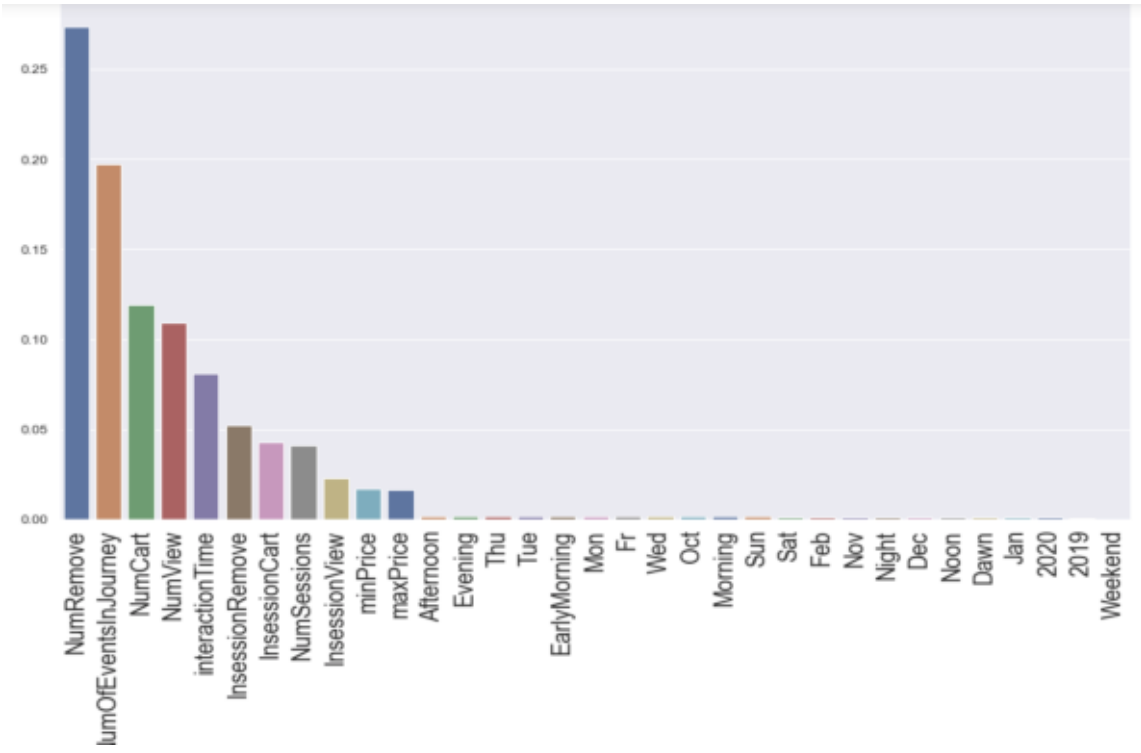

(a)

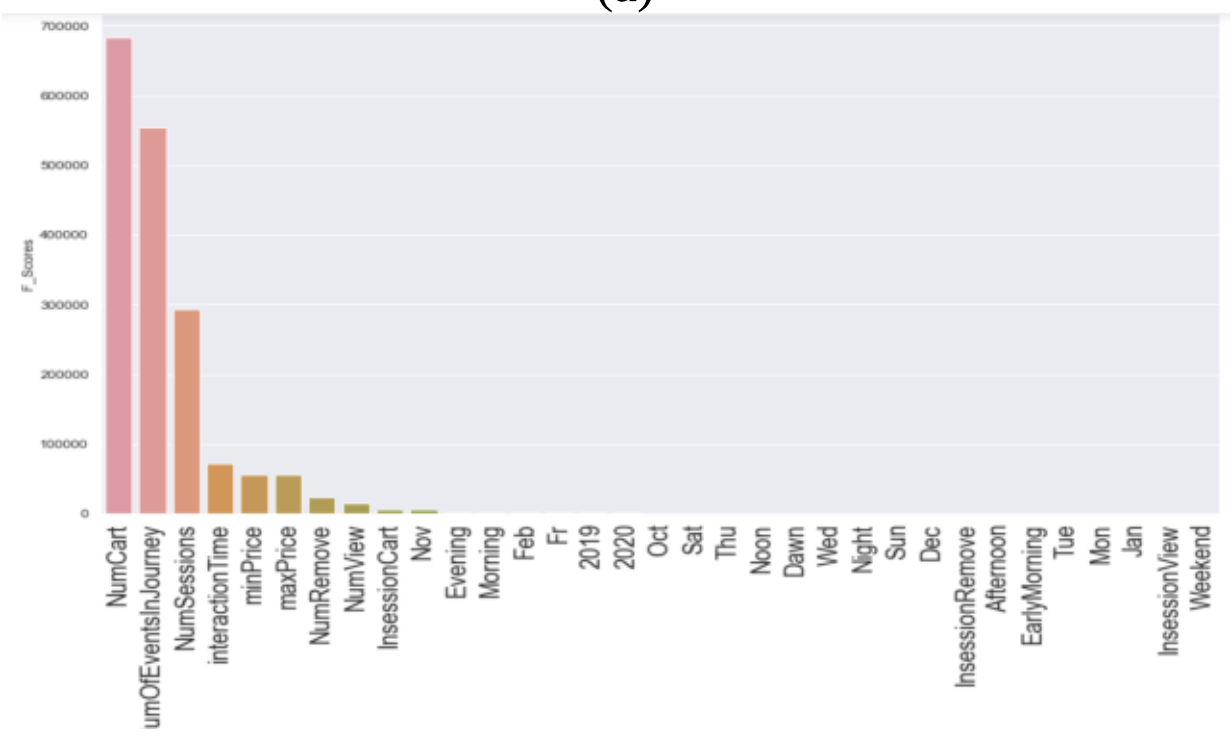

(b)

Fig. 6. Feature Ranking using (a) Random Forest and (b) Fisher Score for Journey Features on Cosmetics data set, respectively..

TABLE II. SELECTED SESSION BASED FEATURES

| | **Dataset** | |
|---|---|---|
| | ***Cosmetics*** | ***Electronics*** |
| Selected Features | TotalEventsInSession<br>NumBrandsCartedInSession<br>NumProdsCartedInSession<br>NumTimesCartedinSession<br>NumTimeRemovedinSession<br>NumTimeViewedinSession,<br>NumBrandsViewedinSession,<br>NumProdsViewedinSession | AvgAmtCartedInSession,<br>NumBrandsCartedInSession<br>NumCategoriesCartedInSession<br>NumProdsCartedInSession,<br>NumTimesCartedinSession,<br>OverallAmtUserCarted,<br>TotalEventsInSession,<br>interactionTime<br>NumBrandsViewedInSession,<br>NumBrandsViewedInSession |

# V. EXPERIMENTS AND RESULTS

We perform three sets of experiments corresponding to the three parallel modules in the proposed system. First, we perform predictive analytics using an optimal set of engineered features to classify journeys and sessions that resulted in classification from the non-purchase counterparts. Second, we implement an LSTM model to predict purchase behaviors on a session level. The probabilistic outcome per session indicates if the next session is likely to result in a purchase or not. Third, we perform unsupervised clustering of user journey-level data to infer customer insights that are necessary to inform marketing and customer retention strategies.

## *A. Classification of User Journey-Data*

For our user journey based data $\{X_J,, Y_J\}$, we implemented binary classification with 70/30 split with and without class imbalance resolving methods (since the data is inherently class imbalance). First, we train data models (Logistic Regression and KNN) on the unbalanced datasets. For the KNN model, the optimal value of neighborhood parameter K is estimated by performing cross validation while varying K as odd numbers in the range 2 to 30.

Next, we perform sample balancing using class weighting and a minor class super-sampling technique called SMOTE [8]. The classification performances on the cosmetics and electronics data sets using the unbalanced and balanced data sets are shown in Table III. Here, we observe that data balancing significantly improves classification precision, recall and accuracy for both Logistic Regression and KNN models. Also, Logistic Regression yields the best classification performance on both data sets. Also, we observe that since the number of purchase samples ($Y_J$ =1) is significantly low in both data sets, our intention is to maximize recall to ensure purchase events are not missed. This enables prediction of quantities for stocking and inventory to ensure customers have access to the products they are interested in. *fp* events result in over stocking that does not significantly harm customer shopping experience.

TABLE III. CLASSIFICATION PERFORMANCE OF USER JOURNEY DATA. BEST VALUES ARE HIGHLIGHTED.

| Model | Metrics | | |
|---|---|---|---|
| | Recall | Accuray | Precision |
| **Logistic Regression** | | | |
| Cosmetics (Unbalanced Data) | 0.4564 | 0.9249 | 0.8538 |
| Cosmetics (Balanced Data) | **0.9951** | **0.9860** | **0.9773** |
| Electronics (Unbalanced Data) | 0.2108 | 0.9752 | 0.6954 |
| Electronics ( Balanced Data) | **0.9952** | **0.9921** | **0.9890** |
| **KNN (k=5)** | | | |
| Cosmetics (Unbalanced) | 0.4732 | 0.9043 | 0.6411 |
| Cosmetics (Balanced Data) | **0.9092** | **0.8767** | **0.8537** |
| Electronics (Unbalanced Data) | 0.2489 | 0.9774 | 0.8241 |
| Electronics ( Balanced Data) | **0.9534** | **0.9207** | **0.8950** |

## B. *Sequence Models for Session-Level Data*

For the session-level data $\{X_{S'}, Y_S\}$, we implement LSTM models with different layer and neuron structures to identify the best LSTM model for both data sets. At a session-level the percentage of sessions that end in purchases are 9.22% and 10.94% for the cosmetics and electronics data sets, respectively. Thus, in the absence of a trained LSTM model, if all sessions were assigned the major non-purchase class ($Y_S = 0$), we would still achieve 90.78% and 89.06% baseline accuracy for the cosmetics and electronics data sets, respectively. The LSTM models are trained on balanced data using SMOTE [8] and a variety of network structures are analyzed with 1-3 layers of bidirectional LSTM layers with 10-40 neurons per layer. This ablation study is presented in Supplementary Materials.

The network structure resulting in the highest Recall is presented in Table IV. Here, we observe that 1 layer of bidirectional LSTM with 40 and 10 neurons are the best network structure for the cosmetics and electronics data sets, respectively. Additionally, by data balancing followed by LSTM training, we achieve higher than baseline classification accuracy and significantly high Recall. However, for the electronics data set, the precision is low, indicating more than 50% of the sessions predicted as purchase events are falsely predicted to end up in a purchase. One reason for the high *fp* rate in this data set is that view events take up almost 95% of all events, and there is no option to remove_from_cart. Thus, a viewing session most often gets falsely predicted as a purchase event. Further, this disparity in session level data recording across data sets necessitates the need for further segmentation of the sessions based on customer category, to reduce *fps*.

TABLE IV. BEST SESSION-LEVEL CLASSIFICATION PERFORMANCES ON THE TWO DATASETS

| **Best Model** | **Metrics** | | | |
|---|---|---|---|---|
| | ***Recall*** | ***Accuracy*** | ***Precision*** | ***F1-Score*** |
| **Cosmetics Dataset** | | | | |
| 1 layer, 40 neurons | 0.9999 | 0.979 | 0.7733 | 0.8721 |
| **Electronics Dataset** | | | | |
| 1 layer, 10 neurons | 0.7344 | 0.916 | 0.4392 | 0.5497 |

LSTM models are largely used to take a sequence as input and probabilistically predict the following sequence outcome [4]. To assess the importance of the session-level features for the LSTM models, we analyze the performance of the proposed Bi-LSTM Feature models with 'baseline LSTM models', that take in the sequence of events in a session and the time spent on each event as sequence input. Thus, for each baseline LSTM model, the session level events are categorized as: {1=view, 2=cart, 3=remove from cart, 4=purchase}. So the input is a sequence of maximum 100 such events e.g. {1,2,1,1,1,1,3,1,,,.....4}, and the time spent on each event, while the output is probability for purchase event (event=4) or not. Compared to Bi-LSTM feature models, the baseline LSTMs don't capture product price and brand-related information, which could be a key factor to impact purchasing decisions. The comparative assessment for the Bi-LSTM feature models and baseline LSTM models for the cosmetics and electronics data sets are shown in Table V. Here, we observe that the proposed Bi-LSTM with features achieves higher accuracy and recall when compared to the baseline LSTM models on both cosmetics and electronics data sets. However, the Baseline LSTM models have higher precision, or lower *fp* rate. Since our intention is to minimize *fns* to ensure session level stock and inventory being sufficient to meet customer demands at all times, *fps* are less detrimental than *fns*, favoring the proposed Bi-LSTM feature models.

TABLE V. COMPARATIVE EVALUATION OF BI-LSTM MODELS. BEST PERFORMANCES ARE HIGHLIGHTED.

| **Model** | **Metrics** | | | |
|---|---|---|---|---|
| | ***Recall*** | ***Accuracy*** | ***Precision*** | ***F1-Score*** |
| **Cosmetics Dataset** | | | | |
| Bi-LSTM Features | **0.9999** | **0.9796** | 0.7733 | **0.8722** |
| Baseline - LSTM | 0.1402 | 0.9206 | **0.9960** | 0.2458 |
| **Electronics Dataset** | | | | |
| Bi-LSTM Features | **0.7344** | **0.9162** | 0.4393 | **0.5497** |
| Baseline-LSTM | 0.258 | 0.9017 | **0.6233** | 0.365 |

### C. *Customer Insights from User-Journey Data*

While predictive analysis/classification of user journeys and sessions are useful to ensure adequate inventory, there is a huge need to categorize shopping behavior based on shopping perusal patterns (frequency or carting, viewing and overall interaction). Many of today's organizations aspire to become data-driven companies. A success factor for companies is the ability to translate insights from data analytics - *to understand how customers think and behave.* With this insight, we can attract and engage new customers by creating a personalized customer experience and retain existing customers [10].

According to Forbes Insights report [11], 78% of 400 marketing leaders either have, or are developing a consumer data platform, where 44% of those reports that a data platform is helping to improve customer loyalty. Further, customer segmentation enables us to differentiate the customer base.

For this analysis we perform unsupervised clustering on the user-journey data with the optimal feature set. First, we apply k-means clustering and Elbow method (using the Yellowbrick Library) to find the optimal number of user journey clusters (k) that minimize distortion score. For both cosmetics and electronics data sets, the optimal number of clusters identified are 5, as shown in Fig.7 below.

Second, we apply k-means with k=5 clusters to tag each user journey with a unique cluster ID in the t-SNE planes. From our feature selection step on user-journey data, we notice that two features such as total interaction time and number of sessions in journey have significantly high importance when compared to other features for discriminating a journey resulting in purchase event. With the high discrepancy in feature importances, we observe that the principal component analysis (PCA) space is less optimal for subspace clustering when compared to the t-SNE space.

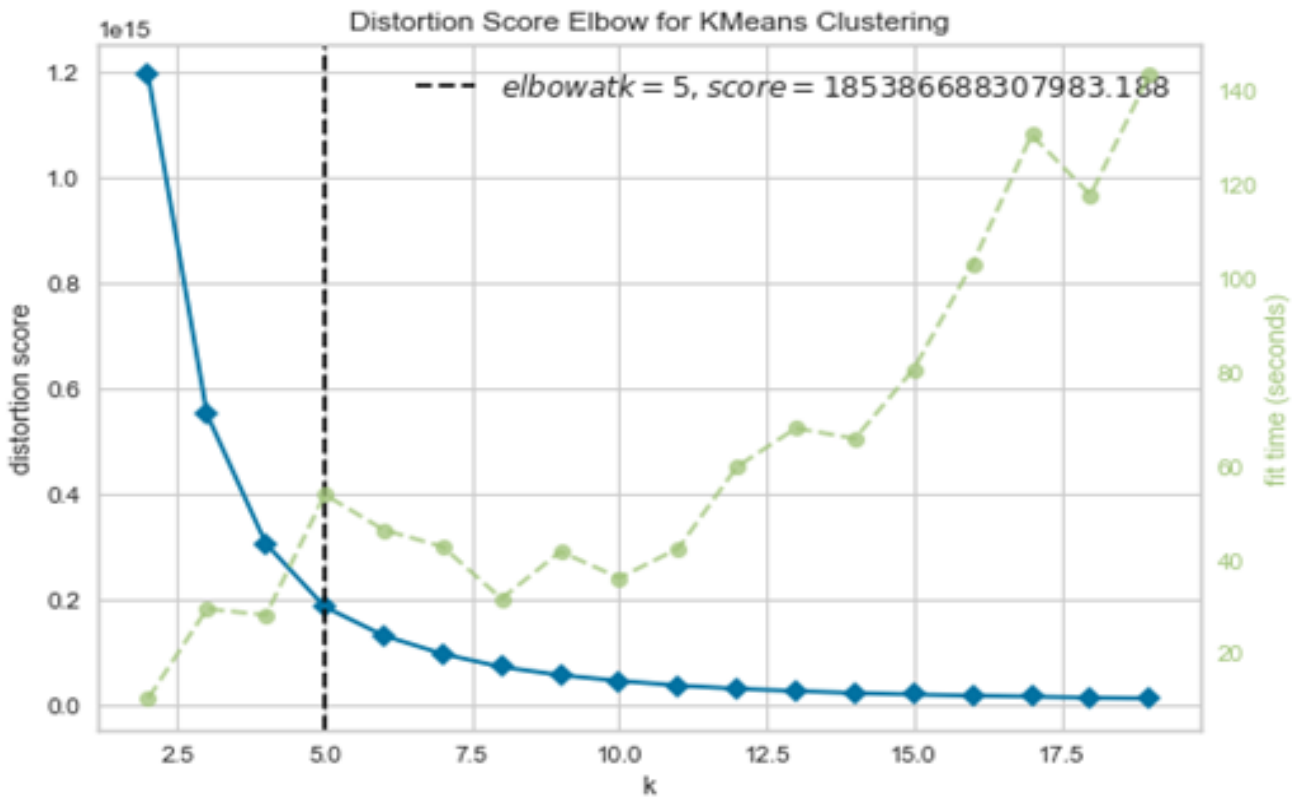

Fig. 7: Elbow method to find k=5 clusters for our data sets.

The representation (Rep) of each cluster (fraction of samples belonging to a particular cluster) and the ratio or purchase journeys in each cluster (PR) are shown in Table VI. We know that at user-journey level the complete Cosmetics and Electronics data sets have purchase ratios around 12% and 1%, respectively. In Table VI, we observe that for both data sets, one major cluster represents 91-99% of the data samples and that cluster has PR *similar* to the PR of the complete data set. Also, as the cluster size decreases, the PR increases upto 3-8 times the overall data PR. Thus, we observe distinct purchasing patterns corresponding to each of the 5 clusters in both data sets that can then be further analyzed for customer insights.

For further assessment of buying behavior corresponding to the 5 different clusters per data set, we analyze the trends along each feature for journeys with purchase events vs journeys with no-purchase events. The goal is to identify features that demonstrate significant variations in purchase vs no-purchase events (e.g. number of purchase samples vs non-purchase are at least 2 times or more) or negative trends (say for all clusters purchase events are higher than non-purchase but for a particular feature and particular cluster purchase events are lower) .

TABLE VI. REPRESENTATION AND PURCHASE RATIO (PR) OF USER JOURNEY CLUSTERS

| | | | | | |
|---|---|---|---|---|---|
| **Cosmetics Data** | | | | | |
| Rep(%) | 91.2 | 4.83 | 2.19 | 1.17 | 0.62 |
| PR(%) | 11.14 | 21.01 | 19.45 | 22.84 | 32.91 |
| Cluster ID | 1 | 4 | 0 | 3 | 2 |
| **Electronics Data** | | | | | |
| Rep(%) | 99 | 0.43 | 0.25 | 0.18 | 0.05 |
| PR(%) | 1.35 | 6.47 | 6.91 | 7.68 | 8.59 |
| Cluster ID | 0 | 3 | 1 | 4 | 2 |

The t-SNE clustering for user journeys from both data sets are shown in Fig. 8. We observe the major class (highest Rep) being visually distinctive from the other smaller classes in the clustering sub-space.

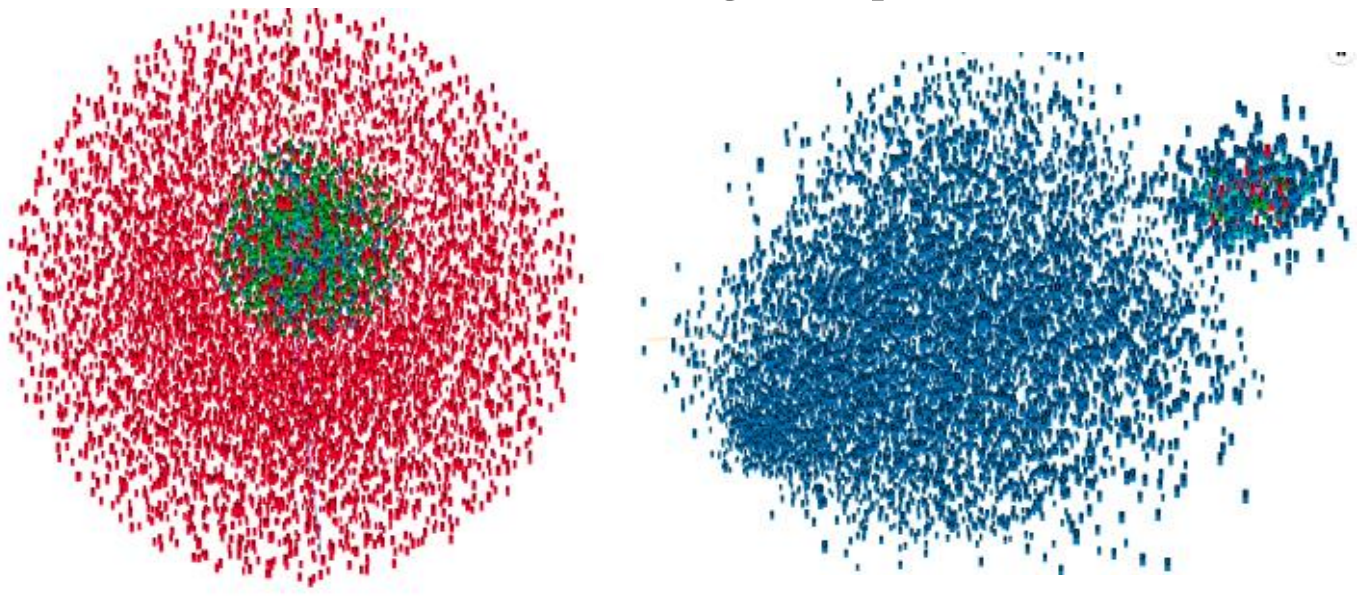

(a) (b)

Fig. 8: The 5 clusters formed for (a) Cosmetics (b) Electronics data sets respectively.

Feature-level analysis of the significant features for purchase vs non-purchase discrimination on each journey cluster is in the Supplementary Material. From this analysis, we find 5 major discriminatory features: total interaction time, number of events and sessions, number of views, average time spent in session views, number of carts and min-max price. Additionally, we observe that for most clusters in both data sets, more views and more time spent on viewing often implies customer indecision resulting in no-purchasing. Also, we observe no significant pattern or trend in separating purchase from no-purchase events for the time-specific features such as time of day, day of week, month etc.

For the cosmetics and electronics data sets the mean values of each feature corresponding to purchase journeys and non-purchase journeys are presented in Fig. 9. Here, the higher feature value is highlighted per cluster. We observe that the major cluster with greater than 90% of all samples, correspond to cluster ID 1 and 0 for the cosmetics and electronics data sets, respectively. This major cluster represents *New Customers/Shoppers* for both datasets with the minimum interaction time, minimum numbers of events and sessions and high price range, indicative of researching intent rather than purchasing intent. Next, we observe that the cluster with highest PR and smallest Rep in both datasets has cluster ID 2 in both data sets. This cluster represents the *Decisive Returning Shoppers* who have the highest interactions, events, sessions, tight price range, less viewing and more carting action indicative of high purchasing intent. Description of the similar cluster types (based on feature level analysis) and the dissimilar cluster types across the two data sets are presented below.

*i. New Shoppers*: Customers that are more likely to look around and find out what they like than purchase. Distinctive features include lowest interaction times, least numbers of sessions, no apparent patterns in viewing, carting and removal actions.

*ii. Impulsive/inquisitive shoppers*: Customers that continue to research but are conscious towards their price range. Distinctive features include second lowest interactions (1-2 sessions, 1-2 carts), slightly more carting behavior than a new shopper with variable viewing and less removing behaviors.

*iii. Intentional/Decisive Shoppers:* Customers that have completed preliminary research, have some idea of what they want in a particular price range. Distinctive features include moderately high interactions, higher carting action (more carts per session and less viewing) than the previous two groups, and less per session views and removals.

*iv. Returning Decisive Shoppers:* Customers that are highly engaged and driven to purchase rather than look around. Distinctive features include highest interaction time, most sessions, carts, and less viewing and removing actions.

The one dissimilarity in the clusters formed for the two data sets are as follows:

*v. The brand shoppers:* Customers for the electronics data set that are highly brand conscious and prefer higher priced brands. Distinctive features include high interaction time, high prices, with low viewing and removal actions. This group represents the luxury group, as the customers are brand sensitive but educated enough to follow through with a purchase event.

*vi. The educated perusing shoppers:* Customers for the cosmetics data set who know their budget and know some products but are keen to research more. Distinctive features include moderate interaction times, more than average sessions, more viewing and less carting action.

For both the data sets, we conclude that the recommendation for marketing and business intelligence units would be to try to increase interactions for *New*

*Shoppers*, increase promotions for *impulsive and perusing shoppers* since they have higher than average PR, and to retain the highly *engaged decisiv*e and *brand sensitive shoppers*, respectively.

Segmentation of customer base using the proposed methods is essential to the personalization that contemporary consumer demands. Research from Infosys [9] shows that 31% of surveyed consumers wish their shopping experiences were far more personalized than it currently is. Given the findings above, we can personalize the online shopping experiences from the different customer categories separately to drive up engagement that will subsequently lead to higher PR.

## VI. Conclusions and Discussions

In this work we present an analytical framework that performs predictive modeling on session-levels and user-product interaction journey levels using automated modules for feature engineering and feature selection. The novelty of this work is in the user-journey based analysis and for predicting purchase events and for categorizing customers based on their purchasing patterns. Additionally, we present scalable models and steps to work around the computational complexities of datasets that are 2GB (cosmetics dataset) to 15GB (electronics dataset) in size.

From our experiments we make three key conclusions. First, session level prediction of purchase events has lower accuracy (91-97%) and precision (43-77%) when compared to user-journey-based classification (97-99% accuracy and precision) for purchase. This is intuitive since repeated user-product interaction information (from user journeys) better represent purchasing intent than any particular session. Thus, prediction of purchasing events at a journey-level is more accurate for stocking and inventory.

Second, Bi-LSTM models with session-level features have superior session-level purchase event classification performance when compared to sequence LSTM models, since the prior captures more information regarding price and brand sensitivities. We observe 1-5% improvement in session-level classification accuracy and 50-70% improvement in classification recall by using session level features as LSTM inputs over the sequence LSTM models.

Third, we extend our analysis to cluster user-journey data with the intention to identify patterns in purchasing journeys. We observe five distinctive clusters emerge for both data sets and further analysis of feature trends to discriminate purchase vs no-purchase events aids distinctive customer insights. While majority of the user-journeys (>90%) belong to clusters representing *New Shoppers* with higher tendencies to research than purchase, there are other minority clusters that demonstrate varying degrees of engagement and purchasing intent. This analysis aids targeted recommendation and follow ups in the form of marketing nudges/coupons/offers for an improved personalized online shopping behavior.

Future work can be directed towards sequence modeling at cluster level and extension of the proposed framework to larger and more diverse data sets.

| Cluster | 0 | 0 | 1 | 1 | 2 | 2 | 3 | 3 | 4 | 4 |
|---|---|---|---|---|---|---|---|---|---|---|
| Feature | Purchase | No-Purchase | Purchase | No-Purchase | Purchase | No-Purchase | Purchase | No-Purchase | Purchase | No-Purchase |
| NumOfEventsInJourney | **6.07** | 4.06 | **3.06** | 1.65 | **8.82** | 5.30 | **6.90** | 4.75 | **4.74** | 3.72 |
| NumSessions | **3.41** | 2.68 | **1.62** | 1.17 | **4.28** | 3.43 | **3.73** | 3.15 | **2.96** | 2.47 |
| interactionTime | **3131000.67** | 3113476.60 | **57566.96** | 21003.86 | **9553337.46** | 9210984.91 | **5898022.39** | 5767501.65 | 1220936.41 | **1244861.33** |
| maxPrice | 5.63 | **8.12** | 4.82 | **10.53** | 6.08 | **8.39** | 5.63 | **7.81** | 5.27 | **8.09** |
| minPrice | 5.45 | **7.86** | 4.80 | **10.52** | 5.74 | **7.98** | 5.39 | **7.48** | 5.14 | **7.93** |
| NumCart | **2.27** | 1.07 | **1.23** | 0.40 | **4.05** | 1.35 | **2.78** | 1.16 | **1.64** | 0.98 |
| NumView | **1.79** | 1.58 | 0.64 | 0.92 | 1.73 | **2.23** | 1.76 | **2.06** | 1.50 | 1.45 |
| NumRemove | 0.84 | **1.40** | 0.17 | 0.33 | 1.35 | **1.72** | 1.03 | **1.53** | 0.52 | **1.29** |
| InsessionCart | 27.99 | **42.43** | **21.76** | 15.70 | 39.49 | **46.24** | 29.74 | **33.63** | 24.06 | **26.66** |
| InsessionView | 20.99 | **29.33** | 13.15 | 13.48 | 25.13 | **28.50** | 19.27 | **22.28** | 18.28 | 18.34 |
| InsessionRemove | 24.88 | **51.46** | 13.41 | 13.73 | 30.28 | **74.92** | 29.35 | **65.31** | 21.91 | **38.88** |
| Cluster Definition for Purchase | • Third highest interaction time (Interaction time>3.1M)<br>• High events in journey (around 6)<br>• Tight budget range (5-8$)<br>• Less than 3.5 sessions<br>• Moderate carting (>2 carts, >27 per session carting time)<br>• High viewing<br>• Less removing | | • Lowest interaction time (Interaction time>50k)<br>• very Low events in journey (3)<br>• High range of budget (4-10$)<br>• More than 1 session<br>• Low (>1.2 carts, >15-21 per sesison carting time)<br>• No patterns in view, remove | | • Highest Interaction time (Interaction time>9.2M)<br>• High events in journey (8)<br>• Tight budget range (5-8$)<br>• High sessions (3-4)<br>• High carting (Per session carting time>39)<br>• Less viewing<br>• Less removing | | • Second highest interaction time(Interaction time>5.7M)<br>• Moderate events in journey (>6)<br>• Tight budget (5-7$)<br>• Sessions>3.5<br>• Average carting (>2.5 carts, >29 per session carting time)<br>• Less viewing<br>• Less removing | | • Interaction time>1.2M<br>• Low events in journey (4)<br>• Tight range of budget (5-8$)<br>• >2.5 sessions<br>• Low carting (>1.5 carts, >24-26 per sesison carting time)<br>• No patterns in view<br>• Less removing | |
| Purchasing Customer Portfolio | **Educated Perusing shopper**<br>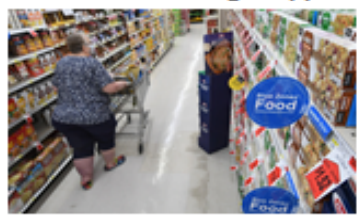 | | **New Shopper**<br>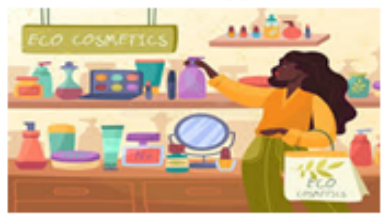 | | **Returning shopper on budget**<br>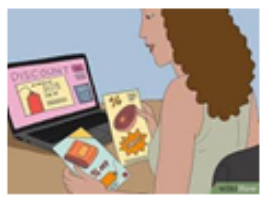 | | **Intentional shopper**<br>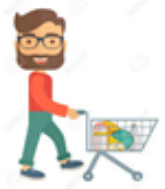 | | **Impulsive perusing shopper**<br>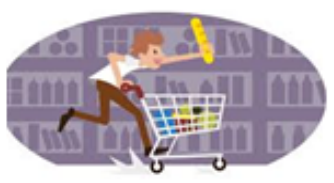 | |

(a)

| cluster | 0 | 0 | 1 | 1 | 2 | 2 | 3 | 3 | 4 | 4 |
|---|---|---|---|---|---|---|---|---|---|---|
| Feature | Purchase | No-Purchase | Purchase | No-Purchase | Purchase | No-Purchase | Purchase | No-Purchase | Purchase | No-Purchase |
| NumOfEventsInJourney | 1.12 | 1.02 | 2.45 | 2.19 | **3.38** | 2.52 | 2.34 | 2.14 | 2.51 | 2.25 |
| NumSessions | 1.05 | 1.01 | 2.34 | 2.14 | **3.19** | 2.49 | 2.22 | 2.09 | 2.43 | 2.20 |
| interactionTime | **1616.85** | 322.08 | 613161.96 | **619363.21** | 1717789.69 | **1791705.77** | **263301.37** | 259325.00 | **1083521.22** | 1072000.57 |
| maxPrice | 297.02 | 290.75 | 375.40 | 365.57 | 327.61 | **382.29** | 326.72 | **348.29** | **403.12** | 383.82 |
| minPrice | 296.91 | 290.74 | 366.20 | 357.11 | 317.63 | **361.65** | 321.95 | **342.17** | **389.92** | 372.05 |
| NumCart | 0.03 | 0.04 | **0.39** | 0.20 | 0.56 | **0.14** | **0.32** | 0.20 | **0.42** | 0.19 |
| NumView | 0.08 | **0.98** | 0.95 | **1.98** | 1.40 | **2.38** | 0.90 | **1.94** | 0.94 | **2.06** |
| InsessionCart | 0.04 | 0.05 | **0.33** | 0.18 | **0.40** | 0.12 | **0.33** | 0.19 | **0.40** | 0.18 |
| InsessionView | 0.15 | **1.21** | 0.83 | **1.27** | 0.90 | **1.27** | 0.85 | **1.31** | 0.84 | **1.30** |
| timeOfDay | 3.47 | 3.72 | 3.33 | 3.59 | 2.90 | **3.63** | 3.02 | 3.43 | 3.49 | 3.60 |
| Cluster Definition for Purchase | • **Lowest interaction time**, less views,<br>• less views per session | | • Marginal interaction time<br>• More carting,<br>• less views,<br>• more carts per session | | • More events in journey,<br>• More sessions,<br>• **High interaction time,**<br>• Lower Price range [317-327]<br>• More carting,<br>• Less views,<br>• More carts per session,<br>• Shop early morning-morning | | • **Moderate interaction time,**<br>• Lower Price range [321-326]<br>• More carting,<br>• Less viewing,<br>• More carts per session | | • **Highest interaction time,**<br>• Higher price range [389-403]<br>• More carting,<br>• Less viewing<br>• More in session carts | |
| Purchasing Customer Portfolio | Lower price points, view less, lowest interactions **(New Inquisitive shopper)**<br>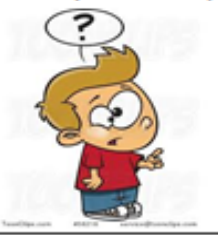 | | Indecisive shopper (similar engagements for purchase/no purchase), but more carting less viewing. **(Impulse shopper)**<br>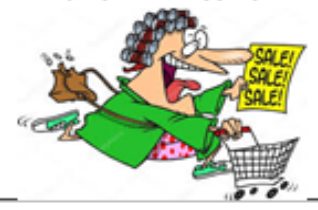 | | Long journeys, pre-informed (more carting less viewing), price sensitive, Decisive. **(Returning Decisive early shopper)**<br>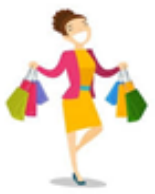 | | Moderate engagement, price sensitive, pre-informed (more carting less viewing), decisive. **(Decisive shopper)**<br>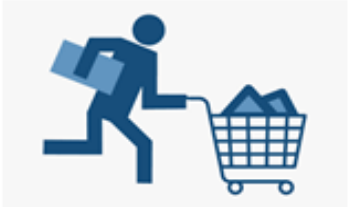 | | High end purchasing, brand sensitive, more carting less viewing, High engagement **(Brand shopper)**<br>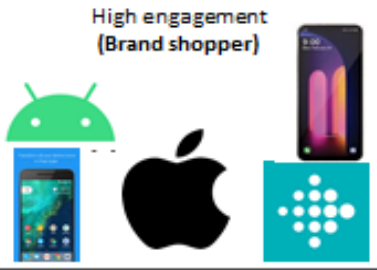 | |

(b)

Fig 9: Customer Insights gathered from the 5 clusters in (a) Cosmetics Data (b) Electronics Data set, respectively.

Supplementary Material to: Categorizing Online Shopping Behavior from Cosmetics to Electronics: An Analytical Framework

## I. Data Description:

Table I below shows a high level comparison of these two datasets.

TABLE I. **DATASET COMPARISON**

| **Property** | **Cosmetics Dataset** | **Multi Category Store Dataset** |
|---|---|---|
| Size | 2.27 GB | 13.67 GB |
| Time range | 08/2019 - 02/2020 | 10/2019 - 11/2019 |
| Number of rows | 20,692,840 | 109,950,743 |
| Number of columns | 9 | 9 |
| Number of event types | 4, one of [view, cart, remove_from_cart, purchase]. | 3, one of [view, cart, purchase]. |
| Missing values (if any) | category_code : 20,339,246 (98%)<br><br>brand : 8,757,117 (42%)<br><br>user_session : 4,598 | category_code : 35,413,780 (32%)<br><br>brand : 15,341,158 (14%)<br><br>user_session : 12 |

The columns in both these datasets are typical to the columns found in a consumer behavior datasets. Table 2 below lists the columns and their descriptions. Each row in a dataset represents an event and all events are related to products and users. However there are different types of events included in the two datasets. While the cosmetics dataset has all four event_types viz. 'view', 'cart', 'remove_from_cart' and 'purchase', the electronics dataset does not include the ' remove_from_cart' event type.

**TABLE II. DATASET COLUMNS**

| Column name | Column description |
| --- | --- |
| event_time | Timestamp when event happened (UTC) |
| event_type | Type of event. One of [view, cart, remove_from_cart, purchase] |
| product_id | Product ID |
| category_id | Product category ID |
| category_code | Category meaningful name (if present) |
| brand | Brand name in lower case (if present) |
| price | Product price |
| user_id | Permanent user ID |
| user_session | User session ID |

*A. Data Cleaning*

*The following data cleaning steps were performed on the two data sets:*

a. Dropping the ‘category_code’ and ‘brand’ column for the cosmetics dataset due to excessive number of null values.
b. Dropping the ‘category_id’ column for the multi-category store dataset because of availability of the more meaningful ‘category_code’ column.
c. Dropping all rows in both the datasets where price is less than 0. There were 131 only such records in the Cosmetics dataset.
d. Dropping rows in both datasets where user_session is missing.
e. Dropping the rows in the Multi-category store dataset where the ‘category_code’ does not include a category related to electronics. For more details on the different values of electronics related ‘category_code’ in the multi-category store dataset (also hereon referenced as electronics dataset in this paper), please see Appendix.
f. Dropping the rows in the Electronics dataset where ‘category_code’ or ‘brand’ has null values.
g. Formatting Column Types : Setting correct column types like datetime for the ‘event_time’ column, float for the ‘price’ column and string for all other columns.

## II. Computational Framework

Using pandas.DataFrame.unstack and pandas.pivot_table instead of apply proves to be much more efficient in terms of execution time as well as simplicity of operations. It must be duly noted that there is no steadfast rule on which function to use amongst apply, unstack or pivot_table and the functions must be employed taking into context the results we are trying to achieve.

For e.g. to calculate the total number of ‘view’, ‘cart’ and ‘remove_from_cart’ events in a given user session the below approaches can be employed out of which the Approach 2 proves to be the most efficient

Approach 1 - Using apply functions to add a new binary column for each event_type to the original dataframe and then grouping the dataframe by user_session and summing over each session to yield the number of ‘view’, ‘cart’, ‘remove_from_cart’ events in each user session.

The execution time for this approach is 0.15 seconds for 1 million records when executed on a machine with 16 GB RAM and increases steeply with more number of records.

Approach 2 - Grouping dataframe by user session and unstacking the value counts of the different event types in the column ‘event_type’ using the unstack function.

The execution time for this approach is 0.03 seconds for 1 million records when executed on a machine with 16 GB RAM and thus is at least 5 times faster than the Approach 1.

Approach 3: Using the pandas.pivot_table function on the dataframe with the 'user_session’ column as index , and aggregating the count of ‘product_id’ (or any column with no null values) across the different event types in the column ‘event_type’.

The execution time for this approach is 0.08 seconds for 1 million records when executed on a machine with 16 GB RAM and while it is 1.8 times faster than apply , it is approximately 2.7 times slower than the Approach 1.

The relative execution time of these 3 functions has been documented above to illustrate that in scenarios wherein it's not possible or easy to use the unstack function, it's better to use the pivot_table function rather than the apply function.

For e.g. to calculate the total amount viewed, carted and removed in a given user session the pivot_table approach is faster and simpler when compared to an approach using the apply or the unstack function.

In the last decade we saw the rise of Big Data platforms. Google BigQuery is one of them, which is a serverless, highly scalable data warehouse that comes with a built-in query engine[1]. BigQuery provides powerful computation power that responds in seconds when processing TBs of data, which makes exploratory analysis and data preparation for machine learning model building faster.

## III. Exploratory Data Analysis (EDA)

The first thing that should be understood is that there are some differences between the two datasets, from which certain market patterns might be learned. According to Table 1, the electronics dataset has  a lower percentage of null values in category and brand. Thus, when feature engineering is performed for the cosmetics dataset, the features related to category and brand are skipped, while for the electronics dataset, features related to these columns are considered as  removing data with null values is sustainable.

i. Price distribution:

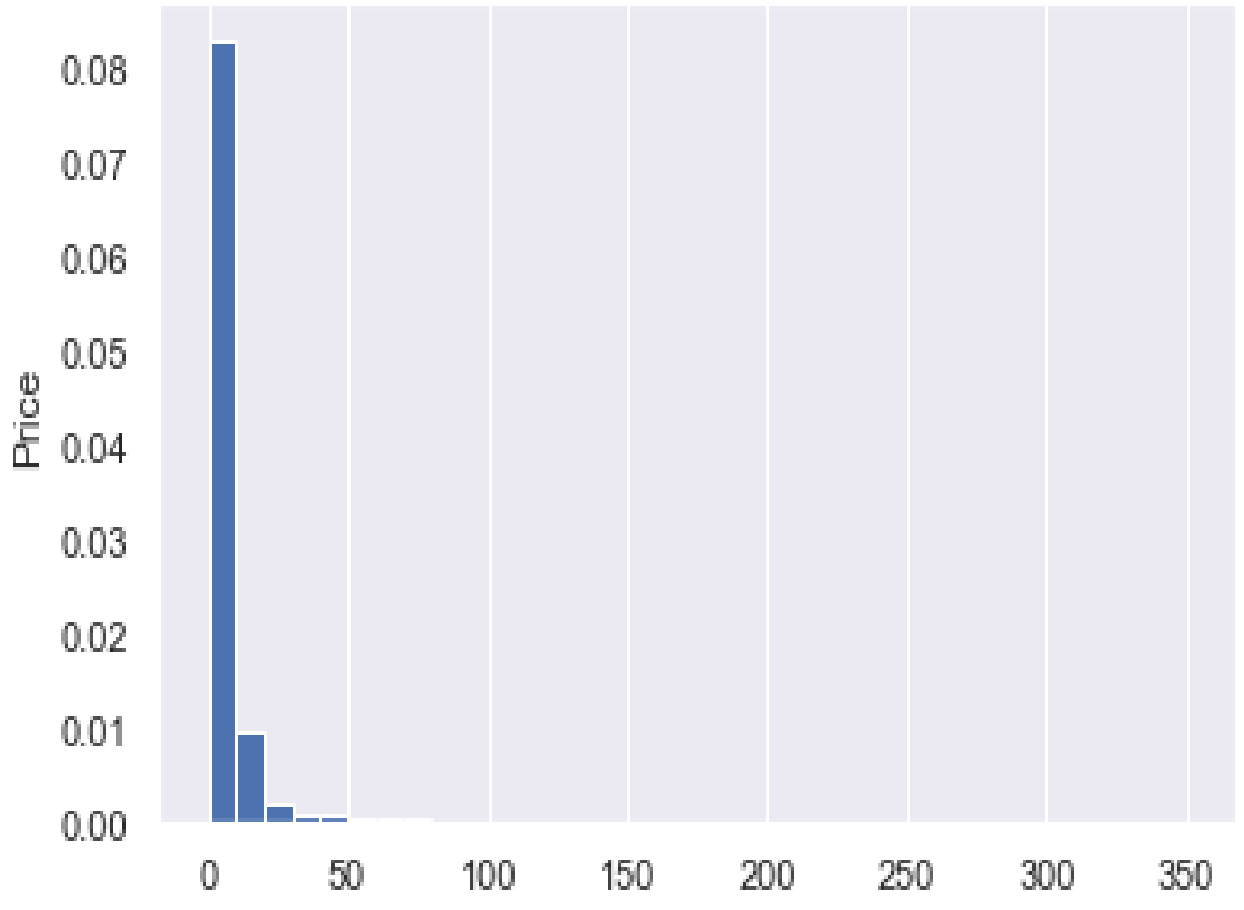

Fig. 1. Price Distribution over Cosmetics Dataset.

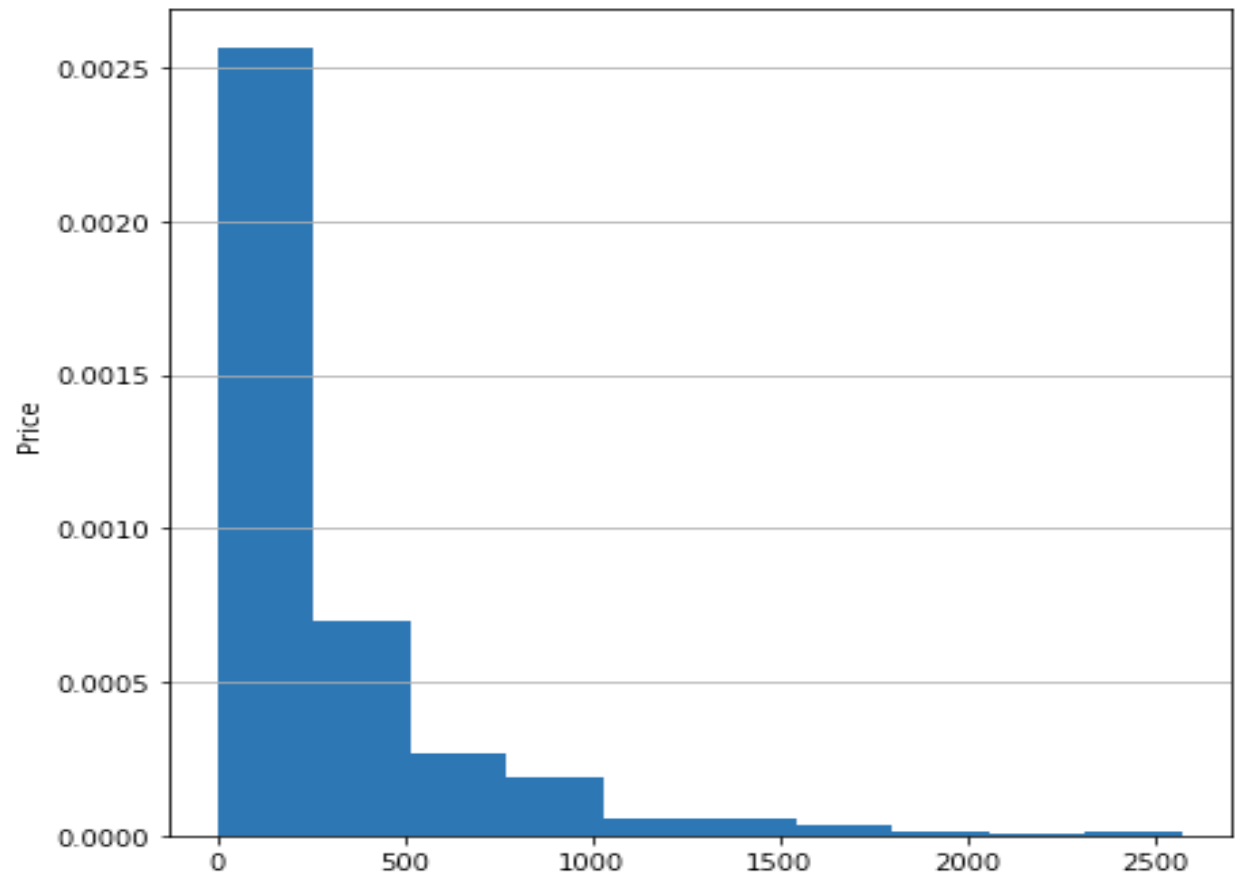

Fig. 2. Price Distribution over Electronics Dataset.

Looking at the price distribution, the mean price of cosmetics products is around 8 dollars, whereas that of electronics products reaches 200 dollars. This variance in price can affect customers' browsing as well as purchasing behaviors.

Take cosmetics data as example, distributions are as follows:

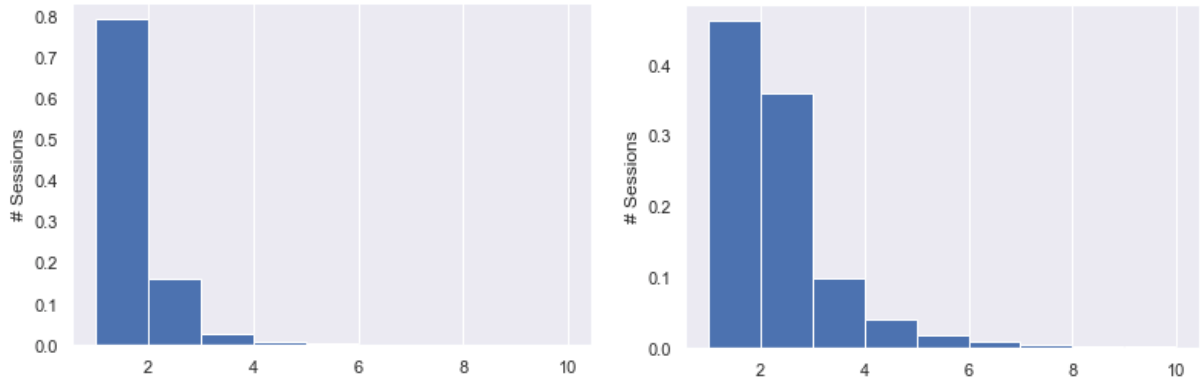

Fig. 1. Distributions of the number of sessions a customer spent on one item, ends up not purchasing & non purchasing respectively.

The event patterns could also separate conversion or not with different distributions:

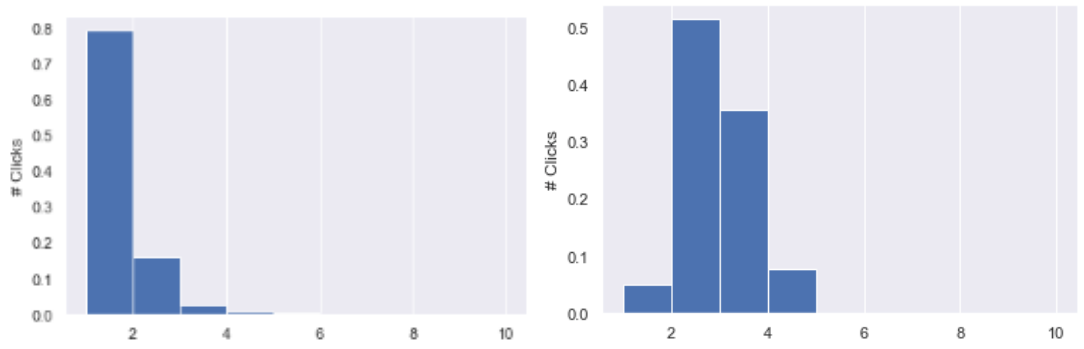

Fig. 2. Distributions of the number of total events a customer creates on one item, ends up not purchasing & non-purchasing respectively.

The number of total events could also be extended to the number of cart events, the number of review events, etc.

iii) Relationship between different event types :

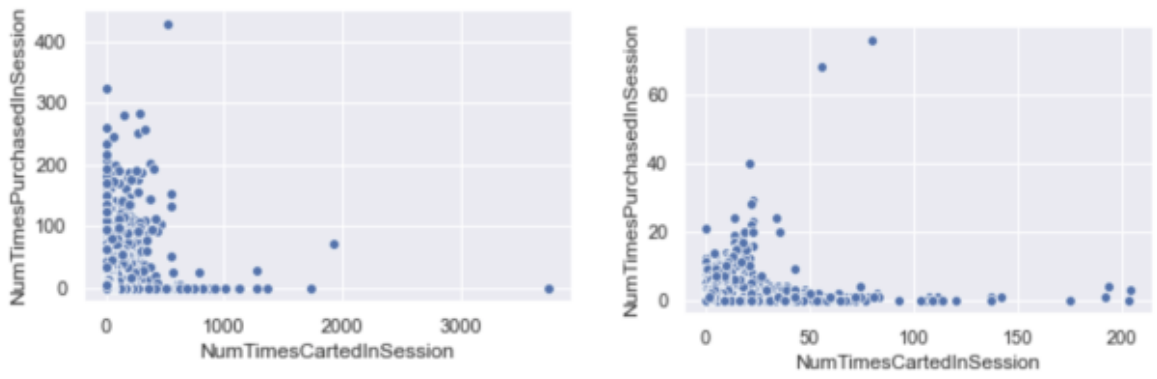

Fig. 3. Relation between number of cart events and number of purchase events in a session for cosmetics & electronics respectively.

iv) Interaction Time

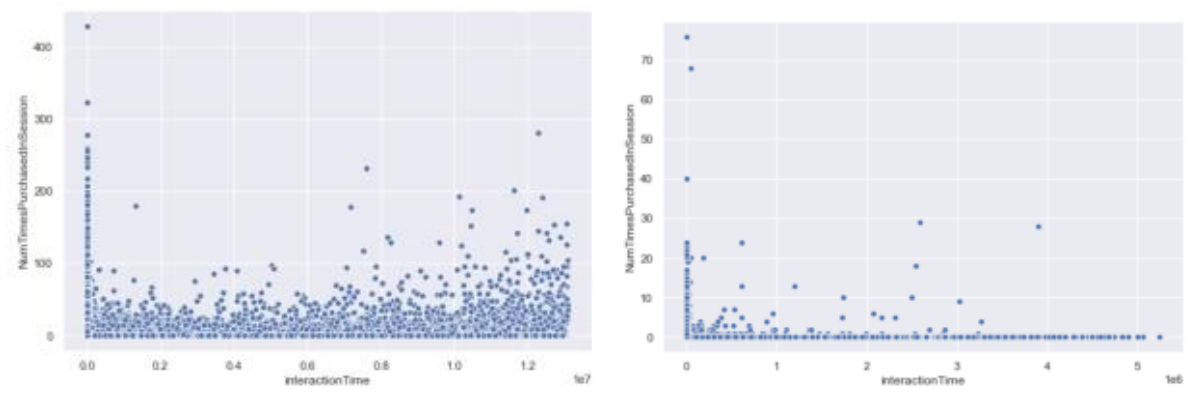

Fig. 4. Relation between interaction time and number of purchase events in a session for cosmetics & electronics respectively.

Inferring from the results of exploratory data analysis, we developed various features on both datasets using different aggregation levels.

*i. User-Journey-based Analysis*

Most e-commerce businesses want to know if a user who is interacting with a certain product on their platform is going to end up buying the product or not. Post exploratory data analysis we realized that both datasets had multiple manifestations of a user's journey. Accordingly, to model a user journey we aggregated data at a user-product level i.e. a given user's all interactions with a single product were bundled together to understand the end outcome of the journey. We then engineered features as shown below in TABLE I. As you may observe the number of features for a user journey are lesser than those for a user session , this is because across a single user journey a user interacts with a single product which belongs to a single brand and a single category and hence there are no features related to these three attributes of a product as there is no variation observed for them across a journey.

TABLE I. **JOURNEY BASED FEATURES**

| **Feature** | **Description** |
|---|---|
| NumOfEventsInJourney | Total number of events in a user journey |
| interactionTime | time difference between the last event and the first event in a journey |
| NumSessions | Total number of sessions in a journey |
| InsessionCart | Maximum amount carted in a session by this user |
| InsessionView | Maximum amount viewed in a session by this user |
| InsessionRemove | Maximum amount removed in a session by this user, only applicable to cosmetics dataset |
| NumCart | Number of times user carted the product in a journey |
| NumRemove | Number of times user removed the product in a journey, only applicable to cosmetics dataset |
| NumView | Number of times user viewed the product in a journey |
| Price | Price of the product |
| Day of week | On which day this journey happens |
| Time of day | At what time of day this journey happens( Early morning, morning, noon, afternoon, dawn, evening, night) |
| Year | Year during which this journey happens |
| Month | Month during which this journey happens |
| Weekend | If the journey happens during the weekend |

Similar to user journey analysis, the data was first aggregated, but here to a session level, and a series of features were digged as follows:

TABLE I. **SESSION BASED FEATURES**

| **Feature** | **Description** |
|---|---|
| TotalEventsInSessio n | Total number of events in session |
| interactionTime | time difference between the last event and the first event in session |
| NumTimesCartedInS ession | number of ‘cart’ events in session |
| NumTimesViewedIn Session | number of ‘view’ events in session |
| NumTimesRemoveIn Session | number of ‘remove_from_cart’ events in session |
| AvgAmtCartedInSes | average amount of prices of |
| sion | carted products in session |
| AvgAmtViewedInSession | average amount of prices of viewed products in session |
| AvgAmtRemoveInSession | average amount of prices of removed products in session |
| NumBrandsCartedInSession | number of different brands carted in session |
| NumBrandsViewedInSession | number of different brands viewed in session |
| NumBrandsRemovedInSession | number of different brands removed in session |
| OverallAmtUserCarted | total prices of the products in the cart |
| OverallAmtUserViewed | total prices of the viewed products |
| OverallAmtUserRemoved | total prices of the removed products |
| NumProdsCartedInSession | number of products in cart in this session |
| NumProdsViewedInSession | number of products viewed in this session |
| NumProdsRemovedInSession | number of products removed in this session |
| Day of week | on which day this session happens |
| Time of day | at what time of day this session happens( Early morning, morning, noon, afternoon, dawn, evening, night) |
| Year | Year during which this session happens |
| Month | Month during which this session happens |
| Weekend | If the session happens during the weekend |

As illustrated earlier the important features were chosen as after performing feature ranking using Random Forest as well as Fisher score and are listed in the table below followed by figures to illustrate the results of feature selection from Random Forests and Fisher Score.

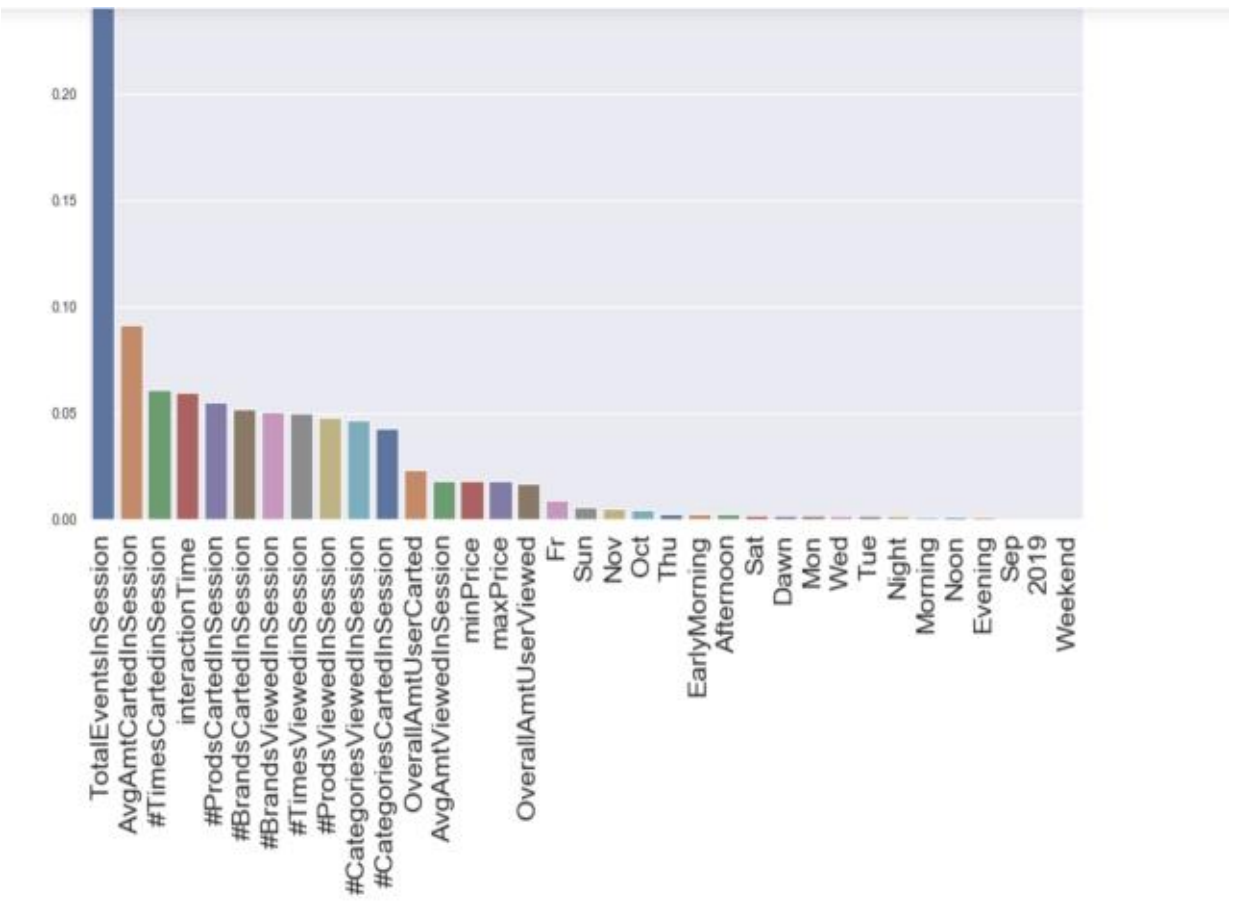

Fig. 1. Feature Ranking as per Random Forests for Session Features.

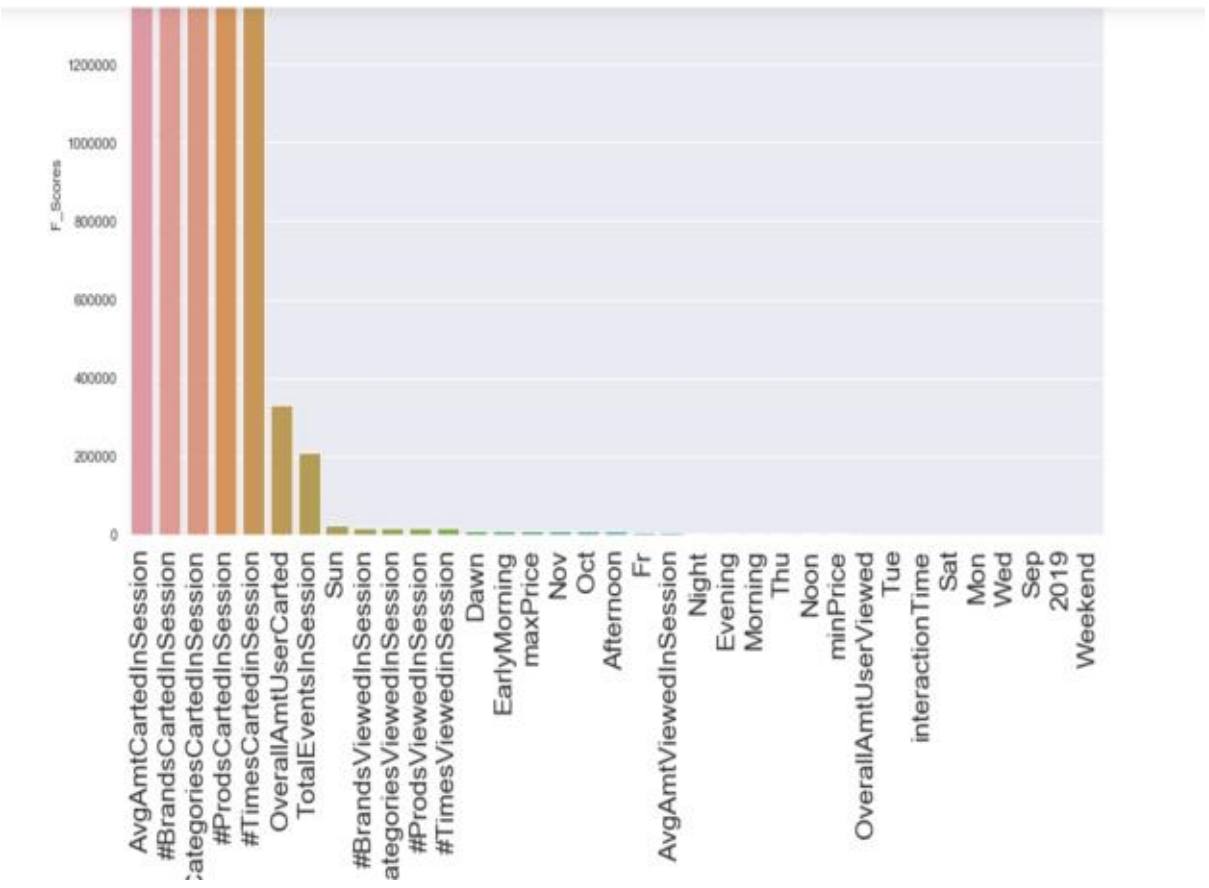

Fig. 2. Feature Ranking as per Fisher Score for Session Features.

## IV. Ablation Study for Sequence Models

*Group 1: Bidirectional LSTM*

*The first group of experiments are meant to compare bidirectional LSTM models with session based features as input data. After determining the optimizer, batch size and number of epochs using cross validation, ablation study is carried out to see the effects of number of LSTM layers and number of output neurons of each LSTM layer on the model performance.*

TABLE I. **BIDIRECTIONAL LSTM ON COSMETICS DATASET**

| **Model** | **Metrics** | | | |
|---|---|---|---|---|
| | ***Recall*** | ***Accuracy*** | ***Precision*** | ***F1-Score*** |
| 1 layer | 0.9999 | 0.9793 | 0.7713 | 0.8709 |
| of 60 neurons | | | | |
| 1 layer of 40 neurons | 0.9999 | 0.9796 | 0.7733 | 0.8721 |
| 1 layer of 20 neurons | 0.9999 | 0.9796 | 0.7733 | 0.8721 |
| 1 layer of 10 neurons | 0.9999 | 0.9796 | 0.7733 | 0.8722 |
| 2 layers 40 neurons + 20 neurons | 0.9999 | 0.9796 | 0.7733 | 0.8722 |
| 2 layers 20 neurons + 10 neurons | 0.2720 | 0.9493 | 0.9998 | 0.4277 |
| 3 layers 40 neurons + 20 neurons+ 10 neurons | 0.4308 | 0.9481 | 0.7096 | 0.5361 |

TABLE II. BIDIRECTIONAL LSTM ON ELECTRONICS DATASET

| **Model** | **Metrics** | | | |
|---|---|---|---|---|
| | ***Recall*** | ***Accuracy*** | ***Precision*** | ***F1-Score*** |
| 1 layer of 60 | 0.4499 | 0.9574 | 0.8789 | 0.5952 |
| neurons | | | | |
| 1 layer of 40 neurons | 0.4502 | 0.9574 | 0.8779 | 0.5952 |
| 1 layer of 20 neurons | 0.4580 | 0.9580 | 0.8823 | 0.6030 |
| 1 layer of 10 neurons | 0.7344 | 0.9162 | 0.4392 | 0.5497 |
| 2 layers 40 neurons + 20 neurons | 0.3763 | 0.9313 | 0.5090 | 0.4327 |
| 2 layers 20 neurons + 10 neurons | 0.4587 | 0.9579 | 0.8790 | 0.6029 |
| 3 layers 40 neurons + 20 neurons+ 10 neurons | 0.7344 | 0.9162 | 0.4393 | 0.5497 |

*improve their sales. Other metrics as accuracy and precision are also listed.*

*Group 2: LSTM vs Others*

*The third group of models includes LSTM, KNN as well as logistics regression.*

TABLE I. **ELECTRONICS DATASET**

| **Model** | **Metrics** | | |
|---|---|---|---|
| | ***Recall*** | ***Accuracy*** | ***Precision*** |
| Bi-LSTM | 0.4502 | 0.95 73 | 0.8779 |
| LR | 0.9952 | 0.99 21 | 0.9890 |
| KNN | 0.9534 | 0.92 07 | 0.8950 |

TABLE II. COSMETICS DATASET

| Model | Metrics | | |
|---|---|---|---|
| | ***Recall*** | ***Acc urac y*** | ***Precisio n*** |
| Bi-LSTM | x | x | x |
| LR | 0.9951 | 0.98 60 | 0.9773 |
| KNN | 0.9092 | 0.87 67 | 0.8537 |

## V. Feature Level Plots for Customer Behavior Clusters

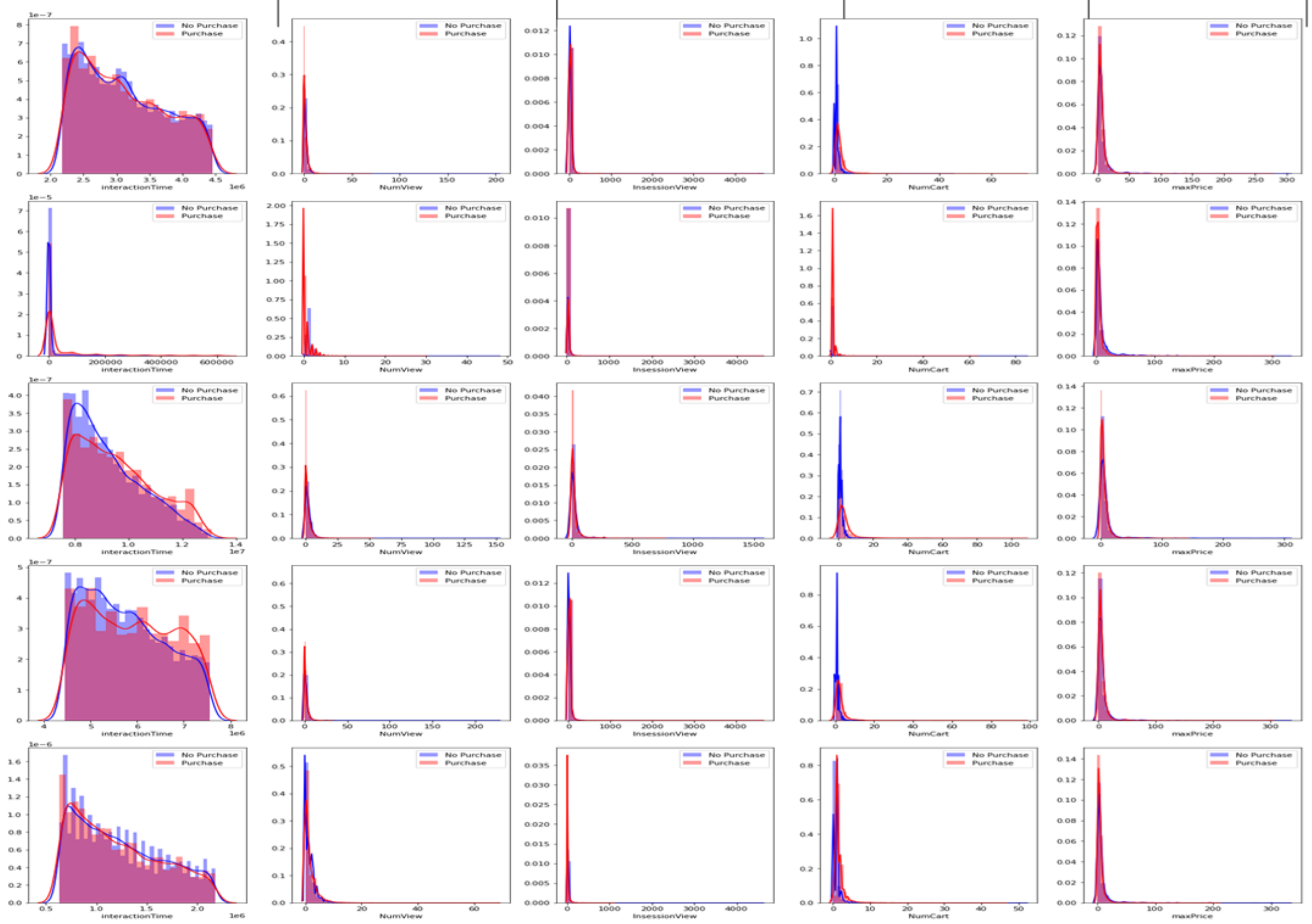

| Cluster | 0 | 0 | 1 | 1 | 2 | 2 | 3 | 3 | 4 | 4 |
|---|---|---|---|---|---|---|---|---|---|---|
| Feature | | | | | | | | | | |
| InteractionTime | | | | | | | | | | |
| NumViews | | | | | | | | | | |
| InsessionViews | | | | | | | | | | |
| NumCart | | | | | | | | | | |
| MaxPrice | | | | | | | | | | |
| Distinction between Purchase vs. No Purchase | Average interaction time, Tight price range, moderate carts per session less viewing. **Educated Perusing shopper** | | Lowest interaction time, High Price range, No pattern in view or remove **New Shopper** | | Highest interaction time, More carting, Less viewing, Tight price sensitive **Returning shopper on budget** | | High interaction time, Tight price range, more carts per session less viewing. **Intentional shopper** | | Low interaction time Tight price range No pattern in view or remove **Impulsive perusing shopper** | |

`Fig. 1: Distribution plots for purchase (Red) vs no-purchase (Blue) customer journeys on the Cosmetics data

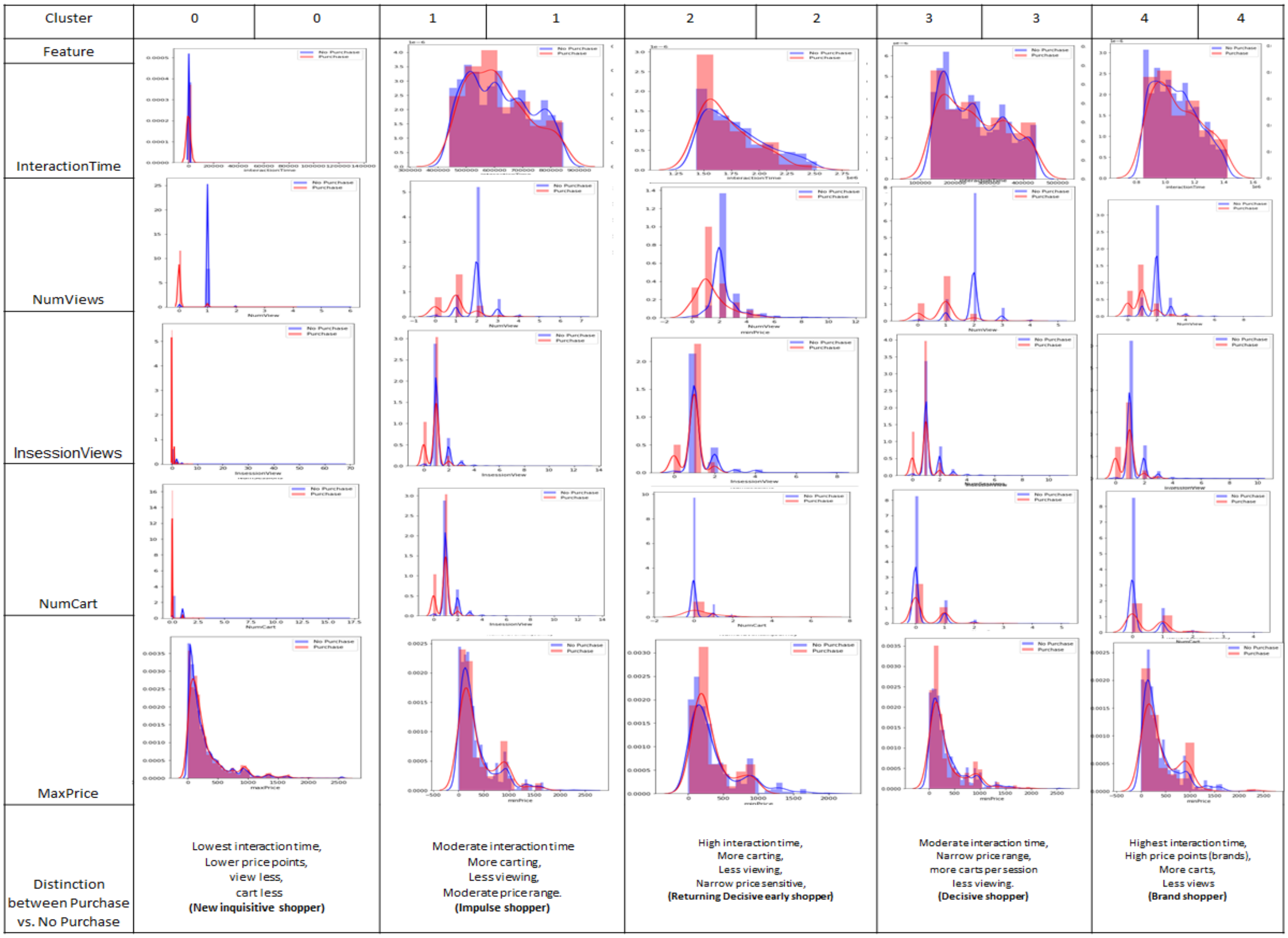

(b)

Fig. 2: Distribution plots for purchase (Red) vs no-purchase (Blue) customer journeys on the Electronics Data set.